\documentclass[runningheads]{llncs}
\usepackage{graphicx}
\usepackage{cite}
\usepackage{amsmath,amssymb,amsfonts}
\usepackage{algorithmic}
\usepackage{graphicx}
\usepackage{textcomp}
\usepackage{xcolor}
\usepackage{multirow}
\usepackage{balance} 
\usepackage{rotating}
\usepackage[utf8]{vietnam}
\usepackage[english]{babel}

\usepackage{array,booktabs}
\newcolumntype{C}{>{\arraybackslash}p{3.8cm}}

\newcommand{\tv}{\textit{Tieq Viet}}

\begin{document}
\title{Understanding Tieq Viet with Deep Learning Models - An Empirical Study
}
%
%
\author{Nguyen Ha Thanh}
%
\authorrunning{Nguyen Ha Thanh}
%
\institute{National Institute of Informatics, Tokyo, Japan 
}
\maketitle              
\begin{abstract}
Deep learning is a powerful approach in recovering lost information as well as harder inverse function computation problems.
When applied in natural language processing, this approach  is essentially making use of context as a mean to recover information through likelihood maximization.
Not long ago, a linguistic study called \tv{} was controversial among both researchers and society. We find this a great example to demonstrate the ability of deep learning models to recover lost information.
In the proposal of \tv{}, some consonants in the standard Vietnamese are replaced. A sentence written in this proposal can be interpreted into multiple sentences in the standard version, with different meanings. 
The hypothesis that we want to test is whether a deep learning model can recover the lost information if we translate the text from Vietnamese to \tv{}.

\end{abstract}
\section{Introduction}
According to Crystal \cite{crystal2011dictionary}, context is to refer to specific parts of an utterance (or text) near or adjacent to a unit which is the focus of attention. The occurrence of a unit is determined by its context, which is specified in terms of the unit’s relations. Thus this definition has confirmed the role of context for placing words in sentences. The word is the basic semantic unit of a sentence so it can be inferred that context has a decisive role in understanding the semantics of a language.

In natural language processing, context can be considered as an important source of information. CoBAn, a context-based model for data leakage prevention \cite{katz2014coban} leveraged strength of both keyword-based and statistical approach in their approach. Automatic detection and correction of spelling are one of the most common problems in natural language processing. A context-based approach like N-gram for Vietnamese Spell Checking \cite{thi2015using} achieve high accuracy approximate 94\% F-score on the Vietnamese text. In research on forum spamming detection, Niu et al. \cite{niu2007quantitative} proposed an effective context-based approach consisting of redirection and cloaking analysis. Recently, deep learning approaches have brought significant breakthoughs in natural language processing \cite{vaswani2017attention,devlin2018bert,brown2020language} in general and Vietnamese text processing \cite{nguyen2018relation,nguyen2019deep,do2021vsec,thanh2021summary} in particular.

In our study, we want to test whether context can help deep learning models in the problem of lost information recovery. The hypothesis we want to verify is: in a situation where most words are modified and the sentence restoring is hampered by ambiguity, can a deep learning model can learn and base on the context to recover the original sentence. 
The experiment is inspired by the controversial work so-called innovating writing of Vietnamese (\tv{}) by Bui Hien \cite{bui2017chu}. This proposal of Vietnamese only contains 31 characters in its character set instead of 38 characters as the standard version. 
As a result, multiple words in the standard version may be translated into the same word in the proposed version, leading to information loss.

\begin{table}
\centering
  \caption{Example for ambiguous in \tv{}. Both translations are valid but convey different (sometimes opposite) meanings.}
  \label{tab:ambiguous_example}
  \begin{tabular}{|C|C|C|}
    \hline
    \textbf{\tv{}}&\textbf{Vietnamese}&\textbf{Meaning}\\
    \hline
    Tôi za cợ mua \textbf{can'}&(1) Tôi ra chợ mua \textbf{chanh} & (1) I went to the market to buy \textbf{lemons}\\
    &(2) Tôi ra chợ mua \textbf{tranh} & (2) I went to the market to buy \textbf{paintings}\\
    
    \hline
    \textbf{Cân câu} ở dây zất nổi tiếq&(1) \textbf{Trân châu} ở đây rất nổi tiếng &(1) \textbf{Pearl} here is very famous\\
    &(2) \textbf{Chân trâu} ở đây rất nổi tiếng & (2) \textbf{Buffalo legs} here are very famous\\
    
    \hline
    Họ tấn kôq zữ zội kuá, ta xôq wể \textbf{xôq} vào& (1) Họ tấn công dữ dội quá, ta không thể \textbf{xông} vào & (1) They attacked so fiercely, we \textbf{must not break in}\\
    & (2) Họ tấn công dữ dội quá, ta không thể \textbf{không} vào & (2) They attacked so fiercely, we \textbf{must enter}\\
    
    \hline
    Kô tìm wấy một kái \textbf{xák} cên bàn làm việk& (1) Cô tìm thấy một cái \textbf{khác} trên bàn làm việc & (1) She found \textbf{another} on her desk\\
    & (2) Cô tìm thấy một cái \textbf{xác} trên bàn làm việc & (2) She found \textbf{a corpse} on her desk\\ 
    
    \hline
    Nó sẽ \textbf{zàn'} một số tiền lớn& (1) Nó sẽ \textbf{giành} một số tiền lớn  & (1) He will \textbf{win} a large amount of money\\
    & (2) Nó sẽ \textbf{dành} một số tiền lớn  & (2) He will \textbf{save} a large amount of money\\
  \hline
\end{tabular}

\end{table}

\section{Problem Formulation}
Consonant phonemes in Vietnamese are represented by single or multiple characters. In \tv{}, existing phonemes are changed as the following:

\begin{itemize}
	\item Ch and Tr are replaced by C
	\item Đ is replaced by D
	\item Gh is replaced by G
	\item Ph is replaced by F
	\item C, and Q are replaced by K
	\item Ng and Ngh are replaced by Q
	\item Kh is replaced by X
	\item Th is replaced by W
	\item D, Gi and R are replaced by Z
	\item Nh is replaced by N'

\end{itemize}
After replacement, the alphabet does not contain the letter Đ any more and is added with some letter such as F, J, W, Z. With such replacement of consonances, the ambiguities between the words are obvious. Some of the ambiguous examples are shown in Table \ref{tab:ambiguous_example}. Without understanding the context, it is impossible to restore the sentences into the standard Vietnamese.

The problem for restoring the sentences from \tv{} to standard Vietnamese can be formulated as follow:
\begin{itemize}
    \item $C_{tieqviet}$ is the character set of \tv{}
    \item $C_{standard}$ is the character set of the standard Vietnamese
    \item Given a sentence $S_{tieqviet}=[c_i|c_i \in C_{tieqviet}]$ obtained by converting a original $S_{standard}=[c_i|c_i \in C_{standard}]$
    \item The system need to restore the sentence as the original version $S_{standard}$.
\end{itemize}

Looking at Table \ref{tab:ambiguous_example}, we can see that there is no restoring rule that can be applied in every reverse conversion. An effective model solving this problem is not a simple mapping between one-to-one from characters in $C_{tieqviet}$ to $C_{standard}$. In the next section, we describe our experiment in restoring sentence from \tv{} and verify the importance of context in language understanding. Given a sentence written in \tv{}, the model needs to restore the original in the standard Vietnamese.

\section{Experiment}
\subsection{Data Preparation}
The data for the experiment is crawled from Vietnamese digital news websites and converted into \tv{} by the rules described in the previous section. The dataset only contains pairs of sentences written in \tv{} and standard Vietnamese. No additional information is added into the dataset.

The training and validation set of data is divided from the dataset. The training set contains 500 pairs of sentences and the validation test contains 100 pairs of sentences.

\begin{table}
  \caption{Parameters of the LSTM model}
  \label{tab:para}
  \begin{center}
  \begin{tabular}{lr}
  \hline
    \textbf{Parameters}& \textbf{Value}\\
    \hline
    Embedding dimension&227\\
    Hidden dimension&512\\
    Number of hidden layers&2\\
    Batch size&32\\
    Sequence length&50\\
    Learning rate&0.001\\
    Number of epochs&300\\
    \hline
    \end{tabular}
   \end{center}
\end{table}

\subsection{Model Architecture}
For input representation for our model, we use character level embedding technique. Each sentence is a sequence of characters $S = [c_1, c_2, ... , c_n]$. In order to enable the model to restore the sentence to the standard version of Vietnamese, we need to obtain a character set $C$ that is the union of the character set of standard Vietnamese and \tv{} ($C_{standard}$ and $C_{tieqviet}$. We also add to $C$ the \textit{UNK} and \textit{PAD} tokens for representing unknown and padding characters. For character embedding, let $m$ is the quantity of set $C$ ($m=|C|$), each character is represented as an one-hot vector $c_i^m$.

In order to process the sentences with different length in a batch, we uniform all sentences into the same length. The maximum sequence length is set to 50 (characters), all sentence longer than 50 characters are pruned and the shorter ones are padded with \textit{PAD} tokens. 

\begin{figure*}
  \centering
  \includegraphics[width=.7\linewidth]{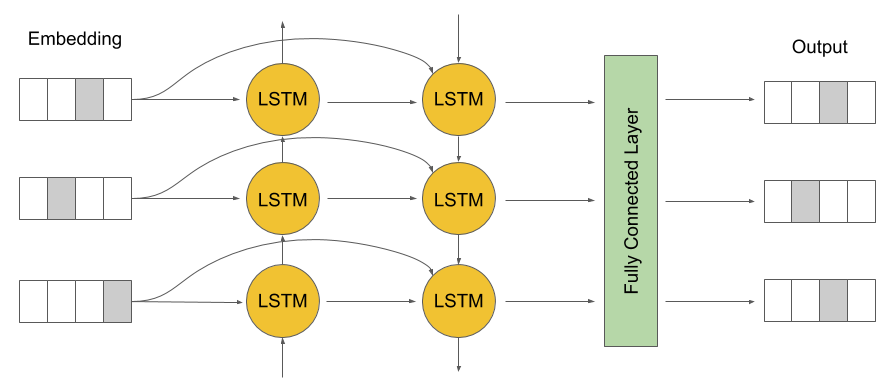}
  \caption{Architecture of the model}
  \label{fig:model_architecture}
\end{figure*}

We use a simple model as 2 stacked bidirectional layers LSTM \cite{sundermeyer2012lstm} followed by a fully connected layer. The architecture of the model is demonstrated in Figure \ref{fig:model_architecture}.

In the training phase, we use Adam optimizer with Cross Entropy Loss to train the neural network with stacked LSTM architecture. The parameters are shown in Table \ref{tab:para}.

\subsection{Experimental Result and Discussion}
\begin{figure}
  \centering
  \includegraphics[width=.8\linewidth]{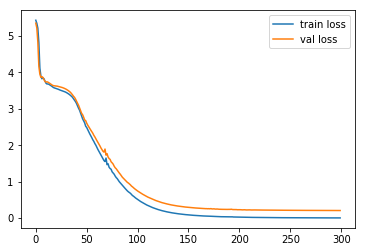}
  \caption{Loss in the training and validation data}
  \label{fig:loss}
\end{figure}

\begin{figure}
  \centering
  \includegraphics[width=.8\linewidth]{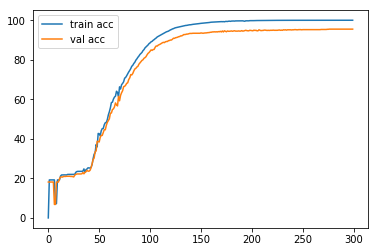}
  \caption{Accuracy in the training and validation data}
  \label{fig:acc}
\end{figure}

After training the model with the parameter listed in the previous section, we obtain a model with 96.32\% accuracy predicting the data in the validation set. The loss and accuracy variation is shown in the Figure \ref{fig:loss} and \ref{fig:acc}. Looking at the graphs, we can see that the model struggle to find the loss optimization direction until epochs 50s.

The loss does not decrease much from epochs 150s. At that period, the model has learned almost restoring rules based on its context understanding and memory. However, the optimization still continues. The less common rules are learned during this period.

The result of prediction in the validation set gives us the quantity measurement for the ability of our model in understanding the context and restoring the text to standard Vietnamese. For further understanding, we investigate cases that the system did not restore successfully. Most of the cases are names, abbreviations or rare words. Table \ref{tab:error} shows some examples of the model prediction errors, $W_{tieqviet}$,$W_{pred}$ and $W_{standard}$ are the word in \tv{}, predicted word by the model and the correct word in standard Vietnamese.

\begin{table}
  \caption{Examples of errors}
  \label{tab:error}
  \begin{center}
  \begin{tabular}{llll}
  \hline
    \textbf{Type}&\textbf{W\textsubscript{tieqviet}}&\textbf{W\textsubscript{pred}}&\textbf{W\textsubscript{standard}}\\
    \hline
    name&Cin'&Trinh&Chinh\\
    name&Tà Zụt&Tà Dụt&Tà Rụt\\
    abbreviation&G&G&GH\\
    name&Cu Zị&Tru Rị&Chu Dị\\
    rare word&cíc&chích&trích\\
    \hline
    \end{tabular}
   \end{center}
\end{table}

\section{Conclusions}
In this paper, we analyze the ability of deep learning models to recover the lost information caused by translating standard Vietnamese into \tv{}.
To remain as generality as possible, we use the simple architecture as LSTM in our experiment.
We also limit the amount of training data and testing data comparatively less than other tasks in NLP.
Our experiment suggests that deep learning models are able to recover the loss of information based on the context with good performance (96.32\%).
This result also provides an empirical fact for scientific debates about \tv{} as well as other Vietnamese linguistic innovation proposals.

\bibliographystyle{splncs04}
\bibliography{ref}

\end{document}